\documentclass[letterpaper, 10 pt, conference]{ieeeconf}  % Comment this line out if you need a4paper

\IEEEoverridecommandlockouts        

\usepackage{graphics} % for pdf, bitmapped graphics files
\usepackage{epsfig} % for postscript graphics files
\usepackage{mathptmx} % assumes new font selection scheme installed
\usepackage{times} % assumes new font selection scheme installed
\usepackage{amsmath} % assumes amsmath package installed
\usepackage{amssymb}  % assumes amsmath package installed
\usepackage{footnote}
\title{\LARGE \bf
Knowledge-aware Graph Transformer for Pedestrian Trajectory Prediction
}

\author{Yu Liu, Yuexin Zhang, Kunming Li, Yongliang Qiao, Stewart Worrall, You-Fu Li, and He Kong
\thanks{This paper was accepted to and presented at the 26th IEEE International Conference on Intelligent Transportation Systems (ITSC), September 2023.}
\thanks{Yu Liu and Yuexin Zhang contributed equally to this work. Yu Liu is with the Shenzhen Key Laboratory of Control Theory and Intelligent Systems, Southern University of Science and Technology (SUSTech), Shenzhen 518055, China and also the Department of Mechanical Engineering, City University of Hong Kong, Hong Kong SAR, China. Email: yuliu254-c@my.cityu.edu.hk. Yuexin Zhang is with the School of Automation, Guangdong Polytechnic Normal University, Guangzhou 510665, China. Email: smileyuexin@163.com. Kunming Li and Stewart Worrall are with the Australian Centre for Field Robotics, The University of Sydney, NSW, 2006, Australia. Email: kuli7393@uni.sydney.edu.au; stewart.worrall@sydney.edu.au. Yongliang Qiao is with the Australian Institute for Machine Learning, The University of Adelaide, SA, 5005, Australia. Email: yongliang.qiao@adelaide.edu.au. You-Fu Li is with the Department of Mechanical Engineering, City University of Hong Kong, Hong Kong SAR, China. Email: meyfli@cityu.edu.hk. He Kong is with the Shenzhen Key Laboratory of Control Theory and Intelligent Systems, SUSTech, Shenzhen
518055, China and also the Guangdong Provincial Key Laboratory of Human-Augmentation and Rehabilitation Robotics in Universities, SUSTech, Shenzhen 518055, China. Email: kongh@sustech.edu.cn. 
        }
}

\begin{document}
\maketitle
\thispagestyle{empty}
\pagestyle{empty}

\begin{abstract}
Predicting pedestrian motion trajectories is crucial for path planning and motion control of autonomous vehicles. Accurately forecasting crowd trajectories is challenging due to the uncertain nature of human motions in different environments. For training, recent deep learning-based prediction approaches mainly utilize information like trajectory history and interactions between pedestrians, among others. This can limit the prediction performance across various scenarios since the discrepancies between training datasets have not been properly incorporated. To overcome this limitation, this paper proposes a graph transformer structure to improve prediction performance, capturing the differences between the various sites and scenarios contained in the datasets. In particular, a self-attention mechanism and a domain adaption module have been designed to improve the generalization ability of the model. Moreover, an additional metric considering cross-dataset sequences is introduced for training and performance evaluation purposes. The proposed framework is validated and compared against existing methods using popular public datasets, i.e., ETH and UCY. Experimental results demonstrate the improved performance of our proposed scheme.

\end{abstract}
%%%%%%%%%%%%%%%%%%%%%%%%%%%%%%%%%%%%%
\section{Introduction}
Pedestrian motion prediction is an essential functionality for self-driving vehicles \cite{ref1}-\cite{ref3}. Given the observed path of pedestrians, accurate predictions of future paths enable autonomous driving vehicles to operate safely and efficiently \cite{ref4}. Pedestrian motions are subject to the effects of several factors such as social and environmental interactions \cite{ref5}.

In this work, we propose a knowledge-aware graph transformer structure to improve the model's adaptability across different scenarios. More specifically, a self-attention mechanism is utilized to capture social interactions, thereby allowing output features to combine temporal and spatial interactions on the motion prediction task. In close spirit to self-attention in the transformer model, a mechanism calculating the affinity relation according to the potential importance of nodes at various sites of the graphs is adopted in the spatial and temporal dimensions, respectively. Attention in the spatial dimension captures interactions between pedestrians dynamically while temporal attention aims to capture the motion patterns of each pedestrian by weighting each point of a trajectory. Knowledge of each node from the spatial and temporal view will be aggregated and shared adaptively into output features. Additionally, a domain adaption mechanism utilizing multiple loss function terms is utilized to reduce the effect of data heterogeneity on the prediction performance. Our contributions can be summarized as follows:

(1) A novel Graph Convolution Network (GCN) based trajectory prediction approach is proposed to capture the motion behaviors of pedestrians, incorporating the heterogeneity of multi-datasets.

(2) A self-attention mechanism and a hybrid loss function are proposed to encode the influence of prediction features. Different from previous aggregation strategies, the approach is more intuitive and effective for adaptively selecting important information from surrounding people. The hybrid loss function for estimating the sample space combines the traditional loss function and the maximum mean discrepancy, and the weights of both are adjustable, thus potentially bringing performance improvements.

(3) Besides the common metrics, a new metric based on the variance of average displacement error is introduced for evaluating the prediction performance. The latter metric can indicate how well the model is able to maintain the prediction robustness under various scenes.

\section{Related Work}
\subsection{Pedestrian Trajectory Prediction}
Due to the interest and developments in autonomous driving, much attention has been paid to trajectory prediction tasks in the literature. Early works on trajectory prediction have adopted various methods, including Gaussian processes \cite{ref6}, Bayesian methods \cite{ref7}, and kinematic and dynamic methods. Other methods \cite{ref8}-\cite{ref9} use physical motion constraints to construct a model of people’s motion tendencies and predict a future path, which has demonstrated a favorable performance. However, in a crowd scene, potential interaction effects between pedestrians is a foreseeable force that will guide future paths and cannot be ignored.

In recent years, deep learning models have become a prevalent tool for the prediction task in a crowd scenario. Social-LSTM \cite{ref10}, as one of the earliest deep learning methods, models the trajectories of each pedestrian using Recurrent Neural Networks (RNNs). Later, some methods \cite{ref11}-\cite{ref13} extend this idea into a generative model. For instance, Social-GAN \cite{ref11} consists of the LSTM-based encoder-decoder architecture; in addition, recurrent neural networks are designed as generators and the generated path will be compared to the ground truth by the discriminator. Another direction is to encourage generalization towards a multi-modal distribution, such as Sophie \cite{ref13} and Social-BiGAT \cite{ref14}. For example, Sophie extracts visual features of scene context via CNN and social features through two way attention mechanism \cite{ref15}, and concatenates the attention output with visual CNN outputs.

To represent the nonregular distribution pattern of pedestrians in a scene and establish their hidden social interactions, graph networks \cite{ref16}-\cite{ref17} have become a popular approach. 
%GAN based approach, Social-BiGAT \cite{ref14}, purposes using GAT \cite{ref9} work to represent the interactions and encourages generalization towards a multi-modal distribution. 
%Graph-based interaction-aware trajectory prediction \cite{ref10} uses a graph to represent the interactions of close objects and a LSTM model to make predictions.
Social-STGCNN \cite{ref17} models every trajectory as graphs in which pedestrians are the nodes and edges are weighted by the adjacent matrices according to the relative distance between the pedestrians to model interaction effects. However, the impact from the temporal view is ignored in such adjacent matrices.   
%Spatio-temporal graph transformer framework \cite{ref12} captures spatio-temporal interaction by a transformer-based graph convolution mechanism.

To sum up, previous works mentioned above have achieved promising progress in trajectory prediction tasks. Nevertheless, existing methods for motion feature extraction in spatio-temporal dimensions and modeling interactions in complex scenes are not sufficient and may lead to poor prediction performance. 

\subsection{Graph Neural Networks}
 To deal with data with non-Euclidean size, graph neural networks (GNN), and their variants are adopted in various cases \cite{ref18}. There are two branches of existing GNN models. One is the spectral GNN method. Data entering the convolution network will be operated on the spectrum domain via Fourier transform. For instance, spectral CNN \cite{ref19} inserts a graph convolution kernel function into each layer on which the convolution operation of nodes is executed. To solve problems of non-local operations in the eigen decomposition process of the Laplacian matrix, ChebyNet \cite{ref20} applies polynomials to replace this process. This method is further developed by graph convolutional networks (GCN) \cite{ref21}, which limits the order of polynomials to just one. 
 
 The other branch is the spatial GNN approach. The convolution operation is directly conducted on the edges of a graph in the spatial domain. For example, GraphSAGE \cite{ref22} aggregates adjacent nodes and fuses nodes in different orders to extract features. In the GAT \cite{ref23} work, an attention mechanism is adopted to assign different importance to nodes in the graph.
 
 Our proposed method is based on the spatial domain approach. However, to fully utilize information of each node, the aggregation strategy will be executed on the spatial and temporal views respectively, through which each node in the graph will adaptively become aware of the existence of other nodes across the spatial and temporal domain.  

\subsection{Interaction Modeling}
Modeling interactions between pedestrians is essential for trajectory prediction. Previous works, such as the social force model \cite{ref9}, rely on hand-crafted features and have demonstrated some effectiveness in predicting crowd motion. 

In recent learning-based methods \cite{ref10},\cite{ref24}, social pooling modules are implemented on the hidden states of RNNs to dynamically capture interactions of people within a certain distance. A similar pooling layer is also adopted in GAN-based approaches such as Social-GAN \cite{ref11}. The interaction feature in Sophie \cite{ref13} is constructed by subtracting the hidden states of LSTM. However, relying on such hidden states to model interactions can not immediately reveal the latest status of the interaction, causing problems when there is a rapid change in orientation over a short time. 

Self-attention, the core idea from the Transformer model \cite{ref26}, establish the affinity relationship between each time frame and has shown some success on the tasks of processing a time sequence. Recent works \cite{ref14}, \cite{ref25} extend the attention mechanism \cite{ref15} to model interactions in spatio-temporal graphs. For example, Social-BIGAT \cite{ref14} adopts a graph structure in the pooling module in which attention networks (GAT) \cite{ref23} formulate pedestrian interactions. 

Different from the work in \cite{ref14}, in our method, every trajectory is modeled as a graph and the relations of nodes are directly formulated in the graph. We extend the multi-head self-attention mechanism into the spatial and temporal dimensions respectively to build relations between each node, aggregating the spatial and temporal information.

\section{APPROACH}
\subsection{Problem Definition}
The task of trajectory prediction is to estimate future location coordinates of all pedestrians walking in a scene based on their past moving trajectories. History positions of all ${N}$ pedestrians in the 2-D spatial coordinates over observed time $t \in{\left\{t_{1}, t_{2}, t_{3}, ..., t_{obs} \right\}} $ are given as $X_t=\left \{X_{t}^{1}, X_{t}^{2}, X_{t}^{3}, ..., X_{t}^{N} \right \} $, where $ X_t^i=\left \{(x_t^i,y_t^i )\right \}$ is the position of the ${i}$th people at time ${t}$. Similarly, the ground truth trajectory of all ${N}$ pedestrians over future time period $t\in{\left\{ t_{obs+1}, t_{obs+2}, t_{obs+3}, ..., t_{pred} \right\}} $ is then donated as $ Y_t=\left \{Y_{t}^1, Y_{t}^2, Y_{t}^3,..., Y_{t}^{N} \right \}$ and $ Y_t^i=\left \{(x_t^i,y_t^i )\right \}$ is the ground truth position of the ${i}$th pedestrian at time ${t}$ in future. The aim of this work is to predict the future trajectories $ \hat{Y}_t=\left \{\hat{Y}_{t}^1, \hat{Y}_{t}^2, \hat{Y}_{t}^3,..., \hat{Y}_{t}^{N} \right \}$ and the likely position $ \hat{Y}_t^i=\left \{(\hat{x}_t^i, \hat{y}_t^i )\right \}$ of the ${i}$th pedestrian at time ${t}$.

%Previous works assume that the pattern of pedestrian trajectories follows a bivariate Gaussian distribution which correlates 2-D spatial coordinates of pedestrians and uses negative log Maximum Likelihood Estimation method as the loss function. In this work,  besides this assumption, we also introduce a domain adaption method into the loss function detailed in the section of loss function.

%As discussed above, previous works have been insufficient in modeling interactions between moving pedestrians. At the same time, the problem that the performance of trained models can not maintain consistency on varying test scenes also is one critical point that cannot be neglected. 

%To mitigate limitations of modeling interactions between moving pedestrians insufficiently and the performance of trained models can not maintain consistency on varying test scenes that previous works are confronting, the proposed method in this work engages a self-attention-based spatial-temporal graph learning and convolution network learning in the time dimension for establishing a bi-variate Gaussian distribution to the predicted coordinates of pedestrian trajectories.

To mitigate the limitations of existing methods in modeling interactions and the issue of performance inconsistency on varying test scenes, the proposed method in this work engages a self-attention-based spatial-temporal graph learning and convolution network learning in the time dimension for establishing a bi-variate Gaussian distribution to the predicted coordinates of pedestrian trajectories.

%To mitigate these limitations, the proposed method in this work engages a self-attention-based spatial-temporal graph learning and convolution network learning in the time dimension for establishing a bi-variate Gaussian distribution to the predicted coordinates of pedestrian trajectories.

\begin{figure*}
\centering
\includegraphics[scale= 0.55]{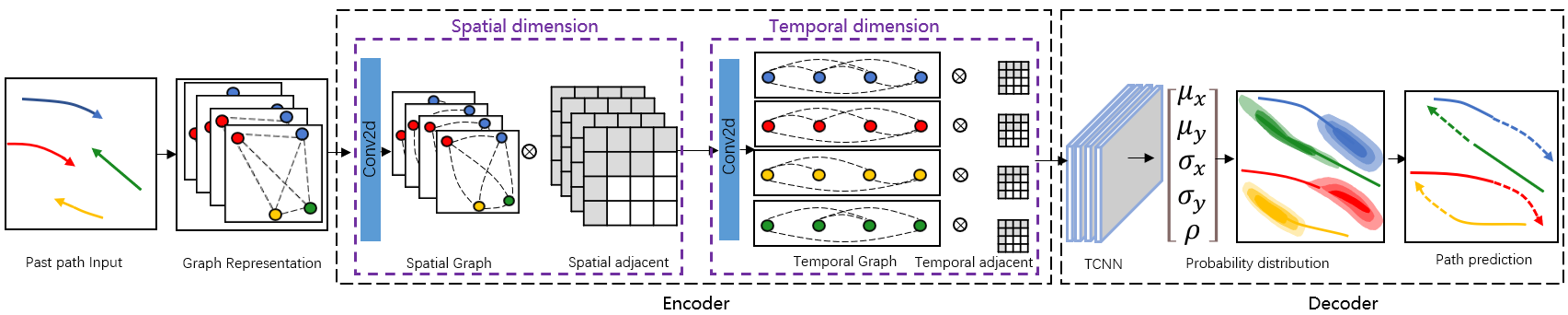}
\caption{The structure of proposed method}
\label{figure structure2}
\end{figure*}

\subsection{The Approach}
The structure of the proposed method is shown in Figure \ref{figure structure2}.
The input frames data will firstly be shaped into the graph representation tensor $X$, including the pedestrians' number, and temporal and spatial dimensions. After that, it will arrive in the spatial graph networks and interaction patterns between nodes will be learned via a self-attention mechanism on the spatial dimension. Then, the motion tendency of each node during each time frame is learned on the temporal dimension using attention-based adjacent matrices, which represent motion relations over past time frames. Finally, the learned path features are fed into a convolution network, which will extract temporal feature patterns and predict parameters of a bi-variate Gaussian distribution for generating possible trajectories. 

%\begin{figure}
%\centering
%\includegraphics[width=3.2in]{figure structure.png}
%\caption{The structure of proposed method}
%\label{figure structure}
%\end{figure}

%\begin{figure*}
%\centering
%\includegraphics[scale= 0.55]{figure Structure3.png}
%\caption{The structure of proposed method}
%\label{figure structure2}
%\end{figure*}

%%%%%%%%%%%%%%%%%%%%%%%%%%%%%%%%%%%%%%%
%that observed position $p_t^n=(x_t^n,y_t^n )$ of pedestrian $n$ at time step $t$ obey a bivariate Gaussian distribution: $p_t^n \sim N(\mu_t^n,\sigma_t^n,\rho_t^n)$. In the same way, the predicted trajectory is represented as $\hat{p}_t^n=(\hat{x}_t^n ,\hat{y}_t^n )$ following the estimated bivariate Gaussian distribution $\hat{p}_t^n\sim N(\hat{\mu}_t^n,\hat{\sigma}_t^n, \hat{\rho}_t^n)$ . Our model is trained to minimize the negative log-likelihood of loss function, which is defined as:
%\begin{equation} \label{equ1}
% L_n=-\sum_{t=1}^n \mathop{log}(P(p_t^n \mid N(\hat \mu_t^n, \hat \sigma_t^n, \hat %\rho_t^n))) 
%\end{equation}
%where $\hat{\mu}_t^n$ is the mean, $\hat{\sigma}_t^n$ is the variance and $\hat{\rho}_t^n$ is the correlation of the distribution.
%%%%%%%%%%%%%%%%%%%%%%%%%%%%%%%%%%%%
\subsection{Spatial Graph Representation}
The trajectory data of pedestrians collected from video frames are in an irregular shape and must first be processed into a regular shape before feeding into the neural network. We first introduce our model by constructing a set of spatial graphs $G_t$ describing the relative positions of ${N}$ pedestrians in a scene at each time $t$. $G_t$ is defined as $G_t=(V_t,E_t)$, in which $V_t=\left\{ {v_t^i \mid \forall i\in \left \{{1,2,3,...,N} \right \} }\right\} $ is the set of vertices $v_t^i$ representing spatial position of each node in the graph $G_t$. $E_t$ is the set of edges of graph $G_t$ and is defined as $E_t=\left \{ {e_t^{ij} \mid \forall {i,j}\in \left \{{1,2,3,..., N} \right \}} \right \}$, where $e_t^{ij}\in \left \{{0,1} \right \}$ indicates whether the nodes ${v_t^i, v_t^j}$ are connected (denoted as 1) or disconnected (denoted as 0), respectively.

However, such an indicator can not provide a detailed relationship between nodes. To explore how strongly the existence of one node nearby could influence the other nodes, an adjacent matrix with weighted coefficients, $a_t^{ij}$, is adopted. Those weighted coefficients, $a_t^{ij}$, are computed for each edge, $e_t^{ij}$, based on the interactions between nodes. Pairs with close distance will obtain larger coefficients, otherwise will be given smaller coefficients. In this case, a multi-head self-attention mechanism from the transformer model is adopted to adaptively train the weighted adjacency matrix $A_t$.
\subsection{Multi-head Graph Attention }
To model the exact interaction between pedestrians involved in the graph, we first adopt the self-attention mechanism to calculate the weighted adjacency matrix at each time step $t$. The input of the spatial attention layer is a set of all feature vectors of all nodes at time step $t$ and defined as $H_t=\left \{{h_n^t \mid \forall n \in \left \{{1,2,3,...,N}\right \} } \right \}$. Similarly, the input vector of the temporal attention layer is defined as $H_n=\left \{{h_n^t \mid \forall t \in \left \{{1,2,3,...,T}\right \} } \right \}$. The multi-head attention mechanism is illustrated as follows:
\begin{equation}
    \begin{array}{cc}
         head_i= Attention(Q,K,V)\\
        MultiHead(Q,K,V)=Concat(head_i,head_2,...,head_h)W^o
    \end{array}
\end{equation}
where $h$ is the counted for the number of attention heads. $Q$, $K$, and $V$ are the query, key, and value vectors, respectively and to gain sufficient power for expressing social interaction. Besides, learnable linear projection matrices, $W_i^Q$, $W_i^K$, $W_i^V$, $W_i^O$, are used:

\begin{equation}\label{equ3}
    \begin{array}{cc}
        Q=W_i^Q H_t, \text{ } K=W_i^K H_t,  \text{ } V=W_i^V H_t
    \end{array}
\end{equation}

 The scaled dot-product attention of attention operation is shown as: 
\begin{equation}\label{equ4}
  Attention(Q,K,V)=softmax(\frac{QK^T}{\sqrt{d_k}})V
\end{equation}
where $d_k$ is the dimension of the key vector. 

\subsection{Spatial Graph Neural Network}
Give input data in the format of tensor $X \in R^{(D \times T \times N)}$, where $N$ denotes the dimension of pedestrians number in an observed scene in $T$ time steps, each node includes coordinate information in ${D}$ dimensions (here ${D}=2$), and each node is represented as $(x_t^n,y_t^n)$. In addition, the input tensor $X_{in}$ will first pass a convolution layer to extract feature vectors denoted as $X^t$, which will expand the dimension of ${D}$ from 2 to 5, the number of parameters in a bivariate Gaussian distribution.

The spatial graph representation of each node and self-attention mechanism is applied to establish a symmetrically weighted adjacency matrix $A_t$ for spatial graph $G_t$ at each time step ${t}$. There are several essential operations before feeding data into the graph layer and implementing graph operation. The symmetrical adjacent matrix $A$ is in the form of $\left \{ {A_1,A_2,A_3,...A_t } \right \}$ and is normalized for each $A_t$ using the following:
\begin{equation}
    A_t=\Lambda_t^{-\frac{1}{2}}\hat{A_t}\Lambda_t^{-\frac{1}{2}}
\end{equation}
where $\hat{A_t}=A_t+I$ and $\Lambda_t$ is the diagonal node degree matrix of $\hat{A_t}$.

Then, the implementation of spatial graph operation is defined as follows:
\begin{equation}
   X_{graph}^t=f(V^t,A_t)=\sigma(\Lambda^{-\frac{1}{2}} \hat{A_t} \Lambda^{-\frac{1}{2}} X^t W_t )
\end{equation}
where $W_t$ is the matrix of trainable parameters, $\sigma$ denotes the activation function of the parametric ReLU (PReLU). After applying the spatial graph layer, the feature vector that represents the graph information is obtained as $X_{graph}^t$.

\subsection{Temporal Graph Representation}
The temporal effect of a single pedestrian at consecutive time series is captured by temporal graphs, which investigate how strongly a pedestrian’s position at one specific time frame can impact its position at other time frames. the temporal graph is defined as follows.

$G_n$, the $n$-th pedestrian’s temporal graph, maintains the location data of $n$-th pedestrian at a continuous time sequence and is defined as $G_n=(V_n,E_n)$, in which $V_n=\left \{{v_n^t \mid \forall t \in \left \{{1,2,3,...,T} \right \} } \right \}$ is the set of the location of $n$th pedestrian at each $t$ time frames and $v_n^t=(x_t^n,y_t^n )$. The edge of the graph $G_n$ is defined as $E_n=\left \{{e_{ij}^n \mid \forall {i,j} \in \left \{{1,2,3,...,T} \right \} } \right \}$. The weighted adjacent matrix $A_n$ for each pedestrian is constructed in the same way by the multi-head self-attention mechanism in the temporal dimension.
\subsection{Temporal Graph Neural Network}
Similarly, before feeding into the graph layer, the output feature of the spatial graph, $X_{graph}^t$, will pass a convolution layer and extract feature vectors denoted as  $X^n$. After that, the extracted feature vectors will flow into the temporal graph layer with the weighted adjacent matrix $A_n$ computed through the self-attention mechanism to perform graph operation at the temporal dimension, which is defined as follows:
\begin{equation}
   X_{graph}^n=f(V^n,A_n)=\sigma(\Lambda^{-\frac{1}{2}} \hat{A_n} \Lambda^{-\frac{1}{2}} X^n W_n )
\end{equation}
where the output, $X_{graph}^n$, is a graph feature of the $n$-th pedestrian in the temporal aspect, the symmetrical adjacent matrix $A_n= \Lambda^{-\frac{1}{2}} \hat{A_n} \Lambda^{-\frac{1}{2}}$. $\hat{A_n}=A_n+I $ and $\Lambda_n$ is the diagonal node degree matrix of $\hat{A_n}$. $W_n$ is the matrix of trainable parameters.

\subsection{Time-extrapolator Convolution Neural Network}
The role of the time-extrapolator convolution neural network (TCNN) is to extract temporal features from input graph embedding and then, to predict future steps. The operation of TCNN works directly on the temporal dimension of input embedding and is able to map input time length into another time length. In this work, the features of past trajectories over eight steps will be mapped into twelve steps in the future.

The input graph embedding will be first reshaped into the format of ${(T \times D \times N)}$ from ${(D \times T \times N)}$, where the dimension of time, ${T}$, is the time length of the observed path. In the first layer of TCNN, twelve, the number of output time steps, kernels are processing convolutional operation on eight, the number of input time steps, channels of input graph embedding to produce output features with the format of ${(12 \times D \times N)}$. Then, twelve kernels also will be used in the remaining convolution layer but will work on twelve channels to generate output data with the same shape of ${(12 \times D \times N)}$, which consequently will be reshaped into the format of ${(D \times 12 \times N)}$ before sending it into a mixed bivariate Gaussian distribution loss model.

\subsection{Loss Function}
%that observed position $p_t^n=(x_t^n,y_t^n )$ of pedestrian $n$ at time step $t$ obey a bivariate Gaussian distribution: $p_t^n \sim N(\mu_t^n,\sigma_t^n,\rho_t^n)$. In the same way, the predicted trajectory is represented as $\hat{p}_t^n=(\hat{x}_t^n ,\hat{y}_t^n )$ following the estimated bivariate Gaussian distribution $\hat{p}_t^n\sim N(\hat{\mu}_t^n,\hat{\sigma}_t^n, \hat{\rho}_t^n)$ . 
%Our model is trained to minimize the negative log-likelihood of loss function, which is defined as:
%\begin{equation} \label{equ1}
% L_n=-\sum_{t=1}^n \mathop{log}(P(p_t^n \mid N(\hat \mu_t^n, \hat \sigma_t^n, \hat %\rho_t^n))) 
%\end{equation}
%where $\hat{\mu}_t^n$ is the mean, $\hat{\sigma}_t^n$ is the variance and $\hat{\rho}_t^n$ %is the correlation of the distribution.
In this work, we take the assumption from previous works that the pattern of each trajectory follows a bi-variate Gaussian distribution. That is the observed position $X_t^n=(x_t^n, y_t^n )$ of the pedestrian $n$ at time step $t$ obeys a bivariate Gaussian distribution : $X_t^n \sim N(\mu_t^n, \sigma_t^n, \rho_t^n)$. In the same way, the predicted trajectory is represented as $\hat{Y}_t^n=(\hat{x}_t^n, \hat{y}_t^n )$ following the estimated bivariate Gaussian distribution $\hat{Y}_t^n\sim N(\hat{\mu}_t^n,\hat{\sigma}_t^n, \hat{\rho}_t^n)$ and defined as :
%that is $\hat{p}_t^n(\hat{x}_t^n,\hat{y}_t^n ) \sim N(\hat \mu_t^n,\hat \theta_t^n,\hat \rho_t^n)$  and defined as: 
%\begin{equation}
%    P(\hat{x}_t^n, \hat{y}_t^n)=\frac{1}{2\pi \sigma_1 \sigma_2 \sqrt{1-\rho^2}}e^{-%\frac{z}{2(1-\rho^2)}}
%\end{equation}
\begin{equation}
    P(\hat{x}_t^n, \hat{y}_t^n)=\frac{1}{2\pi \sigma_{\hat{x}_t^n} \sigma_{\hat{y}_t^n} \sqrt{1-\hat{\rho}_t^{n^2}}}e^{-\frac{z}{2(1-{ \hat{\rho}_t^{n^2} })}}
\end{equation}
where :
\begin{equation}
    z = \! \frac{({\hat{x}_t^n} \!-\! \mu_{\hat{x}_t^n})^2}{\sigma_{\hat{x}_t^n}^2} \!+\!  \frac{({\hat{y}_t^n} \!-\! \mu_{\hat{y}_t^n})^2}{\sigma_{\hat{y}_t^n}^2} \!-\! 2\hat{\rho}_t^n\frac{({\hat{x}_t^n} \!-\! \mu_{\hat{x}_t^n})({\hat{y}_t^n}  \!-\!   \mu_{\hat{y}_t^n})}{\sigma_{\hat{x}_t^n}\sigma_{\hat{y}_t^n}} \!
\end{equation}

%\begin{equation}
%    P(x,y)=\frac{1}{2\pi \sigma_1 \sigma_2 \sqrt{1-\rho^2}}e^{\left \{-\frac{1}{2(1-\rho^2)}[\frac{(x-\mu_1)^2}{\sigma_1^2}+\frac{(y-\mu_2)^2}{\sigma_2^2}-2\rho\frac{(x-\mu_1)(y-\mu_2)}{\sigma_1\sigma_2}]\right \}}
%\end{equation}

The parameters, $\hat \mu_t^n$, $\hat \sigma_t^n$, $\hat \rho_t^n$, of the Gaussian distribution will be trained by convolution networks in the decoder section. Then, according to maximum likelihood estimation theory, the task of the model is to minimize negative log-likelihood loss:
\begin{equation} 
 L_{mle}=-\sum_{t=T_{obs+1}}^{T_{pred}} \mathop{log}[P( Y_t^n \mid N(\hat \mu_t^n, \hat \sigma_t^n, \hat \rho_t^n) )] 
\end{equation}

Bi-variate Gaussian distributions assume the space of probability distributions of random variables are in the same normal distribution shape with a diverse set of mean and variance parameters. Such an assumption is capable of releasing distribution patterns for data with low dimensions describing a property with a low degree of complexity. 
However, the flexibility of pedestrians’ movement and the uncertainty of traffic conditions make trajectories sophisticated and such complications cannot be captured simply by a normal distribution. Further, the statistical tool, maximum likelihood estimation, is to find a set of parameters that allow the model to give the same sample results with the highest possibility. This process is achieved by feeding a tremendous amount of data, which will occupy the entire distribution space. But, there are insufficient data points collected in existing datasets to satisfy this target. Consequently, there are discrepancies between real and generated distributions. 

Therefore, an additional loss function, maximum mean discrepancy (MMD), is introduced in this work. The goal of MMD is to find continuous mapping functions in the sample space and through one of the mapping functions, the mean discrepancy of two distribution spaces, ground true data distribution and predicted data distribution, will reach a maximum value. MMD is defined as follows: 
\begin{equation}
   L_{mmd}=MMD[F,X,Y]:=\mathop{max}_{f \in F}(E_x[f(x)]-E_y[f(y)])
\end{equation}
where $X$ distribution is from the observed datasets and $Y$ distribution is from the predicted result. Then, the total loss is a weighted sum of MMD loss and negative log-like hood loss, which is given as:
\begin{equation}
    L=\alpha L_{mmd}+L_{mle}
\end{equation}
where $\alpha$ is a tuning hyperparameter to balance those two learning losses.

\section{Experiments and Results Analysis}

In this work, the proposed model is trained and tested on two public pedestrian trajectory datasets: ETH \cite{ref27} and UCY \cite{ref28}. Two scenes denoted as ETH and HOTEL are included in ETH, while UCY contains three scenes named as ZARA1, ZARA2 and UNIV respectively. The data points of trajectories in both datasets are sampled every 0.4 seconds. To have a fair performance comparison, we follow the same training and evaluation method of previous works that each time sequence with total 20 time frames (8 seconds) is separated into two parts. The first 8 time frames (3.2 seconds) are regarded as observed path data and the following 12 time frames (4.8 seconds) are considered as predicted ground truth data. The network is implemented in the PyTorch framework. The training batch size is set to 128 and is trained with stochastic gradient descent (SGD) optimizer for 250 epochs. The initial learning rate is 0.01 and changed to 0.002 after 150 epochs.  

To evaluate the performance of the proposed method, two standard metrics are first adopted, average displacement error (ADE) \cite{ref29} and final displacement error (FDE) \cite{ref10}. ADE records the average $L_2$ distance error between all trajectory points from the one with best predicted accuracy and from the ground truth path over the entire time frame, and FDE records the $L_2$ distance error between the final trajectory point from the predicted path and that from the ground truth path. The former, ADE, indicates average prediction precision along the route and the latter, FDE, only gives precision at the final destination point. The two metrics are defined as follows:
\begin{equation}
    \begin{array}{cc}
         ADE=\frac{\sum_{n=0}^N \sum_{t=t_{obs}+1}^{t_{pred}}{\parallel{\hat{p}_t^n-p_t^n}} \parallel _2}{N \times (t_{pred}-t_{obs})}  \\
         
         FDE=\frac{\sum_{n=0}^N \parallel {{\hat{p}_t^n-p_t^n}} \parallel _2}{N}, \text{ } t=t_{pred}
    \end{array}
\end{equation}
where $N$ is the total number of trajectories, which also is the aggregated number of pedestrians in all scenes. $t_{obs}$ is time frame of the observing path, while $t_{pred}$ is time frame of the predicted trajectory.  $\hat{p}_t^n$ and $p_t^n$ are predicted trajectory point and ground truth trajectory point for $n$-th pedestrian at $t$ time frame respectively.

Besides those two common metrics, ADE and FDE, we believe an additional evaluation method is necessary, which can indicate how strongly the predicted model is able to maintain its prediction power under various scenes. The model may be trained sufficiently and performs perfectly in certain traffic conditions. On the contrary, it possibly obtains high ADE and FDE scores. In this situation, the variance of ADE also is recorded and compared. The variance of ADE is measuring the variance of the ADE from each predicted point overall trajectories and is defined as follows:
\begin{equation}
    Var(ADE)=\sqrt{\frac{\sum_{n=0}^N \sum_{t=t_{obs}+1}^{t_{pred}}(\parallel \hat{p}_t^n -p_t^n \parallel _2 -ADE)^2}{N \times (t_{pred}-t_{obs})}}
\end{equation}

\subsection{Quantitative Evaluation}

\begin{table*}
\centering
\caption{ADE/FDE performance compared to proposed method} \label{Table 1}
\begin{tabular}{  c c c c c c c c } 
 \hline
 Method & ETH & HOTEL & UNIV & ZARA1 & ZARA2 & AVG & Var\\ 
\hline
Linear \cite{ref10} & 1.33/2.94 & 0.39/0.72 & 0.82/1.59 & 0.62/1.21 & 0.77/1.48 & 0.79/1.59 & 0.096/0.547\\ 
\hline
S-LSTM \cite{ref10} & 1.09/2.35 & 0.79/1.76 & 0.67/1.40 & 0.47/1.00 & 0.56/1.17 & 0.72/1.54 & 0.046/0.231\\ 
\hline
S-GAN \cite{ref11} & 0.87/1.62 & 0.67/1.37 & 0.76/1.52 & 0.35/0.68 & 0.42/0.84 & 0.61/1.21 & 0.0394/0.142\\ 
\hline
GAT   \cite{ref23} & 0.68/1.29 & 0.68/1.40 & 0.57/1.29 & \textbf{0.29}/0.60 & 0.37/0.75 & 0.52/1.07 & 0.0394/0.142\\ 
\hline
S-STGCNN \cite{ref17} & 0.64/\textbf{1.11} & 0.49/0.85 & \textbf{0.44}/\textbf{0.79} & 0.34/0.53 & 0.30/\textbf{0.48} & 0.44/0.75 & 0.0144/0.0524\\ 
\hline
Proposed & \textbf{0.60}/1.11 & \textbf{0.37}/\textbf{0.58} & 0.47/0.80 & 0.33/\textbf{0.52} & \textbf{0.29}/0.55 & \textbf{0.41}/\textbf{0.71} & \textbf{0.0119}/\textbf{0.0492}\\ 
\hline
\end{tabular}
\end{table*}

\begin{table*}[!htbp]
\centering
\caption{Variance performance compared to proposed method} \label{Table 2}

\begin{tabular}{c c c c c c c c} 
 \hline
 Method & ETH & HOTEL & UNIV & ZARA1 & ZARA2 & AVG & VarAll\\ 
\hline
S-GAN & 0.76/1.79 & 0.40/1.06 & 0.39/0.80 & 0.15/0.36 & 0.21/0.53 & 0.382/0.908 & 0.36/0.79\\ 
\hline
S-STGCNN & 0.4569/1.7875 & \textbf{0.1565}/\textbf{0.7170} & 0.167/0.7981 & \textbf{0.0734}/\textbf{0.4074} & \textbf{0.0925}/\textbf{0.4907} & 0.1893/0.8401 & 0.161/0.885\\ 
\hline
Proposed & \textbf{0.4097}/\textbf{1.54226} & 0.1626/0.8384 & \textbf{0.1633}/\textbf{0.785} & 0.0818/0.4464 & 0.0984/0.5068 & \textbf{0.1811}/\textbf{0.8238} & \textbf{0.1545}/\textbf{0.8613}\\ 
 \hline
\end{tabular}
\end{table*}

\begin{figure*}
\centering
\includegraphics[scale= 0.6]{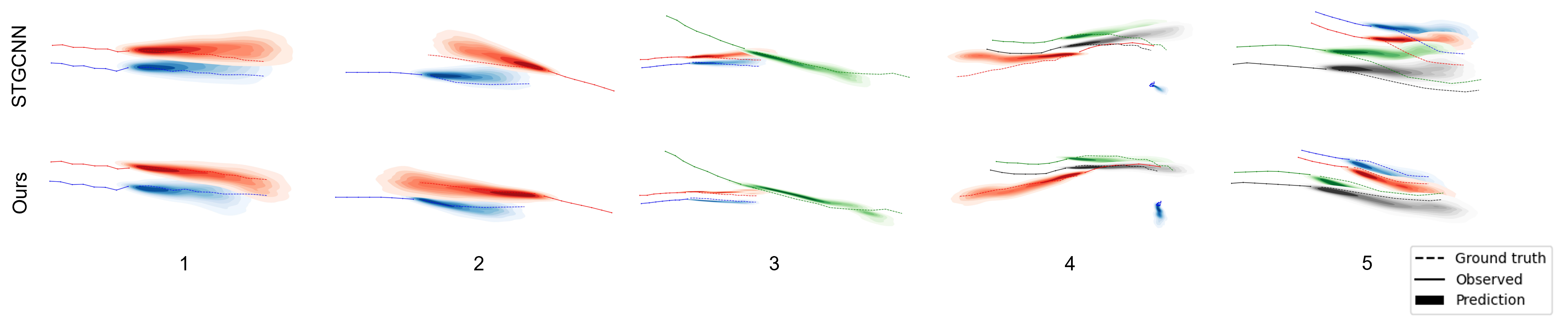}
\caption{Visualization of predicted trajectories distribution across five main scenes. For each example, 300 samples are recorded and their densities are visualized. Observed trajectories are marked as solid lines and the dashed lines are indicating the ground truth future path.  }
\label{Visual1}    
\end{figure*}

\begin{table*}
\centering
\caption{Ablation study of ${\alpha}$ }
\label{Table 3}
\begin{tabular}{c c c c c c c} 
\hline
Method & ETH & HOTEL & UNIV & ZARA1 & ZARA2 & AVG\\
\hline
Base             &        0.64/1.11  & 0.49/0.85 & \textbf{0.44}/0.79 & 0.34/0.53 & 0.30/0.48 & 0.44/0.75 \\
${\alpha}$ = 0.1 &\textbf{0.55}/1.29 & 0.45/0.73 & 0.48/0.87 & \textbf{0.31}/0.62 & 0.39/0.59 & 0.43/0.82 \\
${\alpha}$ = 0.3 &        0.60/\textbf{1.11}  & \textbf{0.37}/\textbf{0.58} & 0.47/0.80 & 0.33/\textbf{0.52} & \textbf{0.29}/0.55 & \textbf{0.41}/\textbf{0.72}   \\
${\alpha}$ = 0.5 &        0.64/1.23  & 0.40/0.63 & 0.46/\textbf{0.74} & 0.37/0.55 & 0.34/\textbf{0.54} & 0.44/0.73 \\
\hline
%\tabnote{ \textbf{Note :} The percentage in the bracket indicates error reduction compared to the Base model. }\\
\end{tabular}
%\footnotetext{This is a text} 
\end{table*}

\begin{table*}
\centering
\caption{The ablation study of each component }
\label{Table 4}
\begin{tabular}{c c c c c c c} 
\hline
Variants & ETH & HOTEL & UNIV & ZARA1 & ZARA2 & AVG\\
\hline
Base             & 0.64/1.11 & 0.49/0.85 & \textbf{0.44}/0.79 & 0.34/0.53 & 0.30/\textbf{0.48} & 0.44/0.75 \\
S-Graph          & \textbf{0.49}/\textbf{1.09} & \textbf{0.36}/0.75 & 0.50/\textbf{0.76} & 0.40/0.52 & 0.44/0.58 & 0.44/0.73 \\
T-Graph          & 0.63/1.23 & 0.39/0.61 & 0.59/0.81 & 0.33/0.60 & \textbf{0.24}/0.56 & 0.43/0.76 \\
ST-Graph         & 0.60/1.11  & 0.37/\textbf{0.58} & 0.47/0.80 & \textbf{0.33}/\textbf{0.52} & 0.29/0.55 & \textbf{0.41}/\textbf{0.72} \\
\hline
%\footnotesize{ \textbf{Note :} The percentage in the bracket indicates error reduction compared to the Base model. }\\
\end{tabular}
%\tabnote{$^{\rm a}$This footnote shows what footnote symbols to use.}
\end{table*}

The proposed method is compared with methods, including, Linear \cite{ref10}, Social LSTM \cite{ref10}, Social GAN \cite{ref11}, GAT \cite{ref23},  and Social STGCNN \cite{ref17}. The test results are shown in Table \ref{Table 1}. We notice that despite continually decreasing prediction discrepancies, a significant variation of ADE and FDE among data sets, which may suggest unstable performance under disparate scenes. In this case, besides ADE and FDE from previous works, the evaluation metrics adopted in this work also include the variance of ADE and FDE over the entire five test data sets.

The results indicate that the average error of our proposed method outperforms all the competing methods on both the ETH and UCY datasets. On the ADE metric, the model overtakes S-STGCNN by 7\% on average for the ETH and UCY datasets. For the FDE metric, the proposed model is better than Social-STGCNN by a margin of 5\% average on the ETH and UCY datasets. The reason behind that is that interactions between pedestrians are dynamically captured on spatial and temporal dimensions. More remarkably, the variance of ADE between tested datasets also has been reduced considerably. Compared to S-STGCNN, our method achieves a decline of 17\% and 6\% on the variance of ADE and FDE respectively.   

The average error variance of the proposed model also outperforms all the previous methods under the introduced metric of variances. As shown in Table \ref{Table 2}, the narrow variance ranges imply that our model maintains its prediction power over different cases. As for the variance of the predicted points in all trajectories in five datasets, our model also keeps the smallest variance.  Hence, the capability of calculating the discrepancy of two distribution spaces by MMD combined with the point estimation method helps the model shape the entire data space.

\subsection{Visualization}

To assess the proposed model visually, several interaction scenes are selected and visualized in Figure \ref{Visual1}, where the solid lines indicate the observed path and dashed lines represent the ground truth path. Different colors correspond to different pedestrians and the depth of color reflects the density of path points estimated by the model. All those scenes are compared with the Social-STGCNN model. The predicted paths in the figure reveal our method is able to capture interactions and give socially acceptable predictions.

To be specific, two pedestrians walking in parallel in the same and opposite directions are selected in scenarios 1 and scenarios 2 respectively. In this relatively simple case where collisions are unlikely to occur, the predicted paths are still on the right track with an acceptable error and relatively small areas for the sample distribution imply better accuracy. Similarly, scenario 3 and scenario 4 show two people walking in a group while the third one is approaching in the same and opposite directions respectively. In more complex cases, such as scenario 5 where four people move forward in parallel, our model closely matches the ground truth.

\subsection{Ablation Study}
In the proposed method, a mixed loss function combining the bivariate Gaussian distribution and a domain adaption method, MMD, is deployed to reduce the discrepancy between data spaces of prediction and ground truth. To further investigate to what extent the balance of these two mechanisms could impact the accuracy of trajectory predictions, an ablation study using different values of ${\alpha}$ is conducted. The results are shown in Table \ref{Table 3}, in which the Base model represents Social-STGCNN. The comparison results suggest that with our introduced domain adaption method, the average error of ADE and FDE has achieved a decrease, which approves the effectiveness of the mixed loss approach. However, such improvements in prediction accuracy are not linearly related as the table shows that the average values of prediction error reach the minimum numbers when ${\alpha}$ equals 0.3 (we will choose this value for ${\alpha}$ in the following).

Similarly, to verify the performance of the attention mechanism on spatial and temporal dimensions, an ablation study is executed. Each test is conducted under the same hyperparameter settings in both the training and testing steps. In this experiment, Social-STGCNN is also used as a base model and the results are shown in Table \ref{Table 4}. In the S-Graph test, the proposed attention mechanism is only deployed in the spatial graph for calculating spatial adjacent matrices without the following graph convolution operation on the temporal dimension before entering TCNN. Reversely, in the T-Graph experiment, the attention mechanism just applies to the temporal dimension for constructing temporal adjacent matrices without the proceeding graph operation on the spatial domain. According to the results, the application of the mechanism has achieved certain effects on two metrics. When integrating both mechanisms, the proposed method further reduces the average error with 7\% and 4\% reduction on ADE and FDE, respectively. 

%\begin{table}[!t]
%\caption{Ablation study of the ${\alpha}$ }
%\label{Table 3}
%\centering
%\begin{tabular}{c c c c c c c}
%\hline
%Method & ETH & HOTEL & UNIV & ZARA1 & ZARA2 & AVG\\
%\hline
%Base             & 0.75/1.35 & 0.47/0.84 & 0.49/0.90 & 0.39/0.62 & 0.34/0.52 & 0.49/0.85 \\
%${\alpha}$ = 0.1 & 0.71/1.29 & 0.45/0.73 & 0.48/0.87 & 0.39/0.62 & 0.37/0.59 & 0.48/0.82 \\
%${\alpha}$ = 0.3 & 0.68/1.30 & 0.43/0.70 & 0.49/0.90 & 0.36/0.58 & 0.33/0.53 & 0.45/0.80 \\
%${\alpha}$ = 0.5 & 0.79/1.33 & 0.40/0.63 & 0.49/0.88 & 0.37/0.57 & 0.34/0.57 & 0.47/0.79 \\
%\hline
%\end{tabular}
%\end{table}

%\begin{figure}
%\centering
%\includegraphics[width=3.2in]{figure 1.png}
%\caption{Predicted trajectory for scenario 1}
%\label{Scenario 1}
%\end{figure}
%
%\begin{figure}
%\centering
%\includegraphics[width=3.2in]{figure 2.png}
%\caption{Predicted trajectory for scenario 2}
%\label{Scenario 2}
%\end{figure}
%
%\begin{figure}
%\centering
%\includegraphics[width=3.2in]{figure 3.png}
%\caption{Predicted trajectory for scenario 3}
%\label{Scenario 3}
%\end{figure}
%
%\begin{figure}
%\centering
%\includegraphics[width=3.2in]{figure 4.png}
%\caption{Predicted trajectory for scenario 4}
%\label{Scenario 4}
%\end{figure}

\section{Conclusion}
We proposed a knowledge-aware graph transformer for pedestrian trajectory prediction, incorporating trajectory information and pedestrian density. The method includes a graph convolution network with the self-attention mechanisms on both spatial and temporal dimensions. A combined loss function mechanism has been designed to predict the trajectories of pedestrians. The graph neural network using self-attention adaptively learns interaction weights among nodes in the scenes and the loss function ensures the shape of sample distribution space. Extensive comparative experimental results illustrate the effectiveness and robustness of our model under various scenes against existing methods. To better mitigate unstable performance under disparate scenes, environmental information such as streets and trees, will be incorporated into the prediction process as future work.  

%As future work, we plan to consider more non-trajectory information (e.g., age, gender and height) to get better prediction.

\section*{ACKNOWLEDGMENT}

This work was supported by the Science, Technology, and Innovation
Commission of Shenzhen Municipality, China [Grant No. ZDSYS20220330161800001].

\end{document}